\documentclass{article}

     \usepackage[preprint]{neurips_2019}

\usepackage[utf8]{inputenc} 
\usepackage[T1]{fontenc}    
\usepackage{hyperref}       
\usepackage{url}            
\usepackage{booktabs}       
\usepackage{amsfonts}       
\usepackage{nicefrac}       
\usepackage{microtype}      
\usepackage{subcaption}

\usepackage{fancyvrb}
\usepackage{pgf}
\usepackage{tikz}
\usetikzlibrary{positioning, arrows,automata}

\usepackage{xcolor}
\usepackage{multirow}
\usepackage{tabularx}
\usepackage{makecell}

\usepackage{amsmath}
\usepackage{amssymb}
\usepackage{amsthm}
\usepackage{pgf}
\usepackage{tikz}
\usepackage{enumitem}

\usepackage{mathtools}

\newtheorem{thm1}{Theorem} 

\DeclarePairedDelimiterX{\setarg}[1]{\{}{\}}{%
  \ifnum\currentgrouptype=16 \else\begingroup\fi
  \activatebar#1
  \ifnum\currentgrouptype=16 \else\endgroup\fi
}

\DeclarePairedDelimiterX{\expectarg}[1]{[}{]}{%
  \ifnum\currentgrouptype=16 \else\begingroup\fi
  \activatebar#1
  \ifnum\currentgrouptype=16 \else\endgroup\fi
}

\newcommand{\prob}{\operatorname{\Pr}\probarg}
\DeclarePairedDelimiterX{\probarg}[1]{(}{)}{%
  \ifnum\currentgrouptype=16 \else\begingroup\fi
  \activatebar#1
  \ifnum\currentgrouptype=16 \else\endgroup\fi
}

\newcommand{\innermid}{\nonscript\;\delimsize\vert\nonscript\;}
\newcommand{\activatebar}{%
  \begingroup\lccode`\~=`\|
  \lowercase{\endgroup\let~}\innermid 
  \mathcode`|=\string"8000
}

\definecolor{colour}{RGB}{219,109,109}
\definecolor{colour2}{RGB}{46, 158, 33}

\title{Learning Portable Representations \\ for High-Level Planning}

\author{
  Steven James and Benjamin Rosman \\
  University of the Witwatersrand \\
  Johannesburg, South Africa\\
  \texttt{\{steven.james, benjamin.rosman1\}@wits.ac.za} \\
  \And
  George Konidaris \\
  Brown University \\
  Providence RI 02912, USA \\
  \texttt{gdk@cs.brown.edu} 
}
\begin{document}

\maketitle

\begin{abstract}
We present a framework for autonomously learning a portable representation that describes a collection of low-level continuous environments.
We show that these abstract representations can be learned in a task-independent egocentric space \textit{specific to the agent} that, when grounded with problem-specific information, are provably sufficient for planning.
We demonstrate transfer in two different domains, where an agent learns a portable, task-independent symbolic vocabulary, as well as rules expressed in that vocabulary, and then learns to instantiate those rules on a per-task basis. 
This reduces the number of samples required to learn a representation of a new task.\vspace*{7mm} 
\end{abstract}

\section{Introduction}

A major goal of artificial intelligence is creating agents capable of acting effectively in a variety of complex environments.  
Robots, in particular, face the difficult task of generating behaviour while sensing and acting in high-dimensional and continuous spaces.
Decision-making at this level is typically infeasible---the robot's innate action space involves directly actuating motors at a high frequency, but it would take thousands of such actuations to accomplish most useful goals.
Similarly, sensors provide high-dimensional signals that are often continuous and noisy. 
Hierarchical reinforcement learning \citep{barto03} tackles this problem  by abstracting away the low-level action space using higher-level \textit{skills}, which can accelerate learning and planning.
Skills alleviate the problem of reasoning over low-level actions, but the state space remains unchanged; efficient planning may also therefore require state space abstraction.
Therefore, we may also wish to perform abstraction in state space.

One approach is to build a state abstraction of the environment that supports planning.
Such representations can then be used as input to task-level planners, which plan using much more compact abstract state descriptors. 
This mitigates the issue of \textit{reward sparsity} and admits solutions to long-horizon tasks, but raises the question of how to build the appropriate abstract representation of a problem.
This is often resolved manually, requiring substantial effort and expertise.

\begin{figure}[h]
	\centering
    \begin{subfigure}[t]{0.45\linewidth}    
			\includegraphics[width=0.7\textwidth]{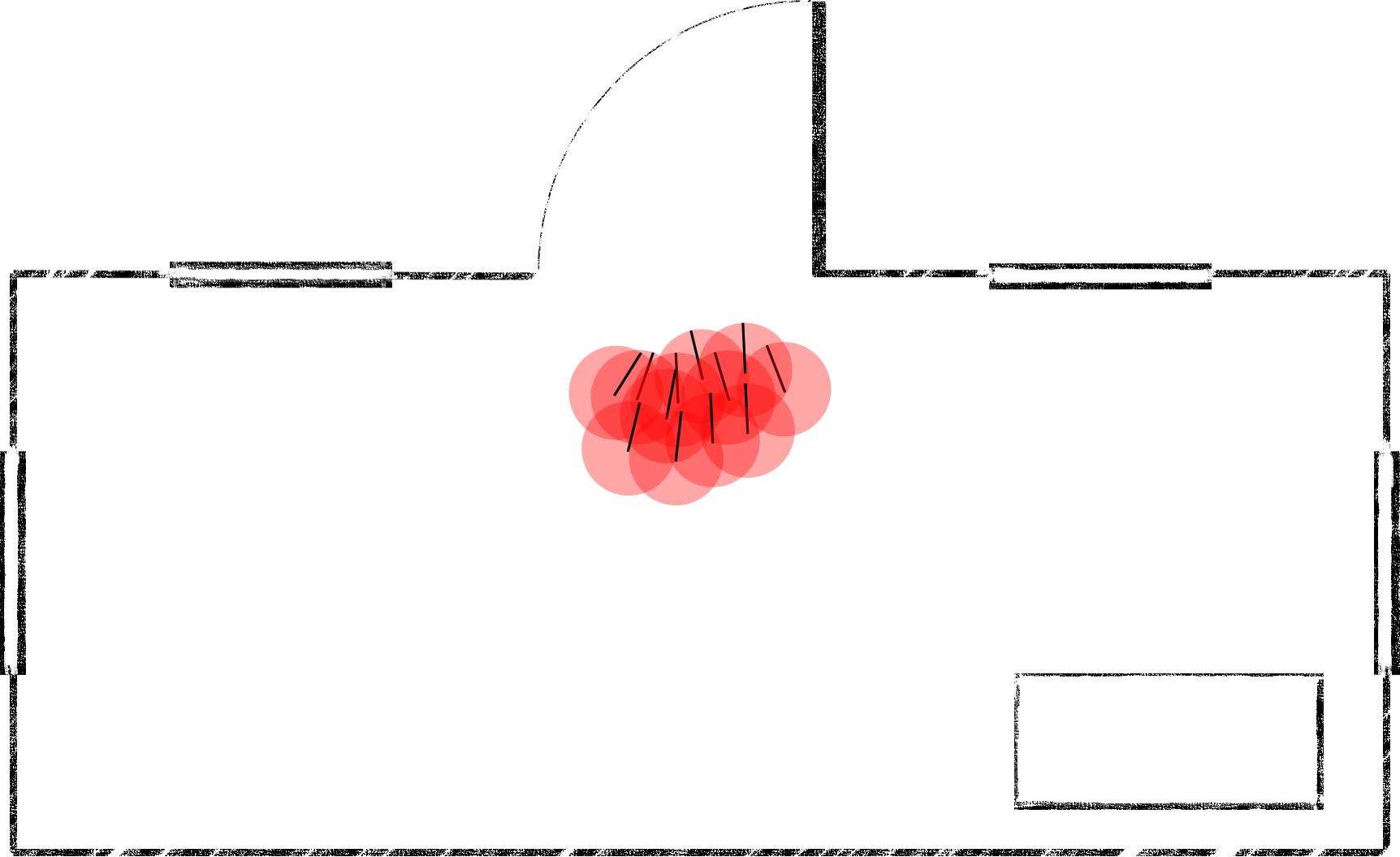}
			 \caption{The distribution over positions from where the agent is able interact with the door. }
			 \label{fig:room1}   
    \end{subfigure}
    \quad
    \begin{subfigure}[t]{0.45\linewidth}    
        \centering
        \includegraphics[width=0.7\textwidth]{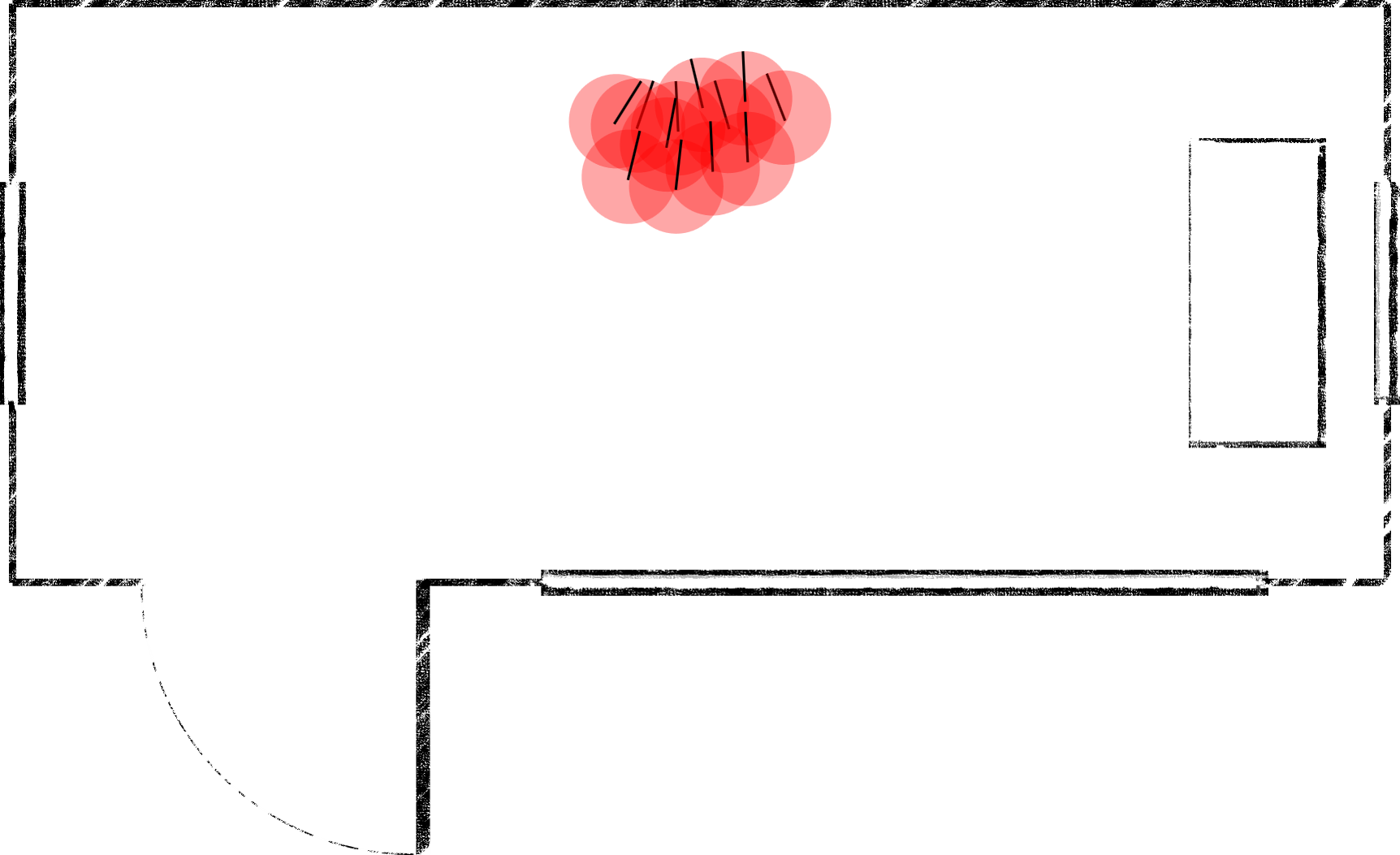}
        \caption{In the new task, the learned distribution is no longer useful since the door's location has changed.}
        \label{fig:room2}
    \end{subfigure}
	\caption{ An illustration of the shortcomings of learning task-specific state abstractions \citep{konidaris18}. (\subref{fig:room1}) An agent (represented by a red circle) learns a distribution over states ($x, y, \theta $ tuples, describing its position in a room) in which it can interact with a door. (\subref{fig:room2}) However, this distribution cannot be reused in a new room with a differing layout.}
	\label{fig:floor}          
\end{figure}

Fortunately, recent work demonstrates how to learn a provably sound symbolic representation autonomously, given only the data obtained by executing the  high-level actions available to the agent \citep{konidaris18}.
A major shortcoming of  that framework is the lack of generalisability---the learned symbols are grounded in the current task, so an agent must relearn the appropriate representation for each new task it encounters (see Figure~\ref{fig:floor}).
This is a data- and computation-intensive procedure involving clustering, probabilistic multi-class classification, and density estimation in high-dimensional spaces, and requires repeated execution of actions within the environment.

The contribution of this work is twofold.
First, we introduce a framework for deriving a symbolic abstraction over an egocentric state space \citep{agre87, guazzelli98, finney02, konidaris12}.\footnote{Egocentric state spaces have also been adopted by recent reinforcement learning frameworks, such as \textit{VizDoom} \citep{kempka16}, \textit{Minecraft} \citep{johnson16} and \textit{Deepmind Lab} \citep{beattie16}.}
Because such state spaces are relative to the agent, they provide a suitable avenue for representation transfer. 
However, these abstractions are necessarily non-Markov, and so are insufficient for planning. 
Our second contribution is thus to prove that the addition of very particular problem-specific information (learned autonomously from the task) to the portable abstractions results in a  representation that \textit{is} sufficient for planning.
This combination of portable abstractions and task-specific information results in lifted action rules that are preserved across tasks, but which have parameters that must be instantiated on a per-task basis.


We describe our framework using a simple toy domain, and then demonstrate successful transfer in two domains.
Our results show that an agent is able to learn symbols that generalise to tasks with different dynamics,
reducing the experience required to learn a representation of a new task. 

\section{Background}

We assume that the tasks faced by an agent can be modelled as a semi-Markov decision process (SMDP) $\mathcal{M} = \langle \mathcal{S}, \mathcal{O}, \mathcal{T}, \mathcal{R} \rangle$, where $\mathcal{S} \subseteq \mathbb{R}^n$ is the $n$-dimensional continuous state space and $\mathcal{O}(s)$ is the set of temporally-extended actions known as \textit{options} available to the agent at state $s$.
The reward function $\mathcal{R}(s, o, \tau, s^\prime)$ specifies the feedback the agent receives from the environment when it executes option $o$ from state $s$ and arrives in state $s^\prime$ after $\tau$ steps.
$\mathcal{T}$ describes the dynamics of the environment, specifying the probability of arriving in state $s^\prime$ after option $o$ is executed from $s$ for $\tau$ timesteps: $\mathcal{T}_{s s^\prime}^o = \prob{s^\prime, \tau | s, o}.$ 
An option $o$ is defined by the tuple $\langle I_o, \pi_o, \beta_o \rangle$, where $I_o = \{ s \mid o \in \mathcal{O}(s) \}$ is the
\textit{initiation set} that specifies the states in which the option can be executed, $\pi_o$ is the \textit{option policy} which specifies the action to execute, and $\beta_o$ is the \textit{termination condition}, where $\beta_o(s)$ is the probability of option $o$ halting in state $s$.

\subsection{Portable Skills}


We assume that tasks are related because they are faced by the same agent \citep{konidaris12}. 
For example, consider a robot equipped with various sensors that is required to perform a number of as yet unspecified tasks.
The only aspect that remains constant across all these tasks is the presence of the robot, and more importantly its sensors, which map the state space $\mathcal{S}$ to a portable, lossy, egocentric observation space $\mathcal{D}$. 
To differentiate, we refer to  $\mathcal{S}$ as \textit{problem space} \citep{konidaris07}.


Augmenting an SMDP with this egocentric data produces the tuple $\mathcal{M}_i = \langle \mathcal{S}_i, \mathcal{O}_i, \mathcal{T}_i, \mathcal{R}_i, \mathcal{D} \rangle$ for each task $i$, where the egocentric observation space $\mathcal{D}$ remains constant across all tasks.
We can use $\mathcal{D}$ to define portable options, whose option policies, initiation sets and termination conditions are all defined egocentrically. 
Because  $\mathcal{D}$ remains constant regardless of the underlying SMDP, these options can be transferred across tasks \citep{konidaris07}.

%
%

\subsection{Abstract Representations}

We wish to learn an abstract representation to facilitate planning.
We define a \textit{probabilistic plan} $p_Z = \{o_1, \ldots, o_n\}$ to be the sequence of options to be executed, starting from some state drawn from distribution $Z$.
It is useful to introduce the notion of a \textit{goal option}, which can only be executed when the agent has reached its goal.  Appending this option to a plan means that the probability of successfully executing a plan is equivalent to the probability of reaching some goal.

A representation suitable for planning must allow us to calculate the probability of a given plan successfully executing to completion.
As a plan is simply a chain of options, it is therefore necessary (and sufficient) to learn when an option can be executed, as well as the outcome of doing so \citep{konidaris18}.
This corresponds to learning the \textit{precondition} $\text{Pre}(o) = \prob{s \in I_o}$, which expresses the probability that option $o$ can be executed at state $s \in \mathcal{S}$, and the \textit{image} $\text{Im}(Z, o)$, which represents the distribution of states an agent may find itself in after executing $o$ from states drawn from distribution $Z$.
An illustration of this is provided in the supplementary material.

In general, we cannot model the image for an arbitrary option;
however, we can do so for a subclass known as \textit{subgoal options} \citep{precup00}, whose terminating states are independent of their starting states \citep{konidaris18}. That is, for any subgoal option $o$, $\prob{s^\prime | s, o} = \prob{s^\prime | o}$.
We can thus substitute the option's image for its \textit{effect}: $\text{Eff}(o) = \text{Im}(Z, o)\ \forall Z$.

Subgoal options are not overly restrictive, since they refer to options that drive an agent to some set of states with high reliability, which is a common occurrence in robotics owing to the use of closed-loop controllers.
Nonetheless, it is likely an option may not be subgoal.
It is often possible, however, to \textit{partition} an option's initiation set into a finite number of subsets, so that it possesses the subgoal property when initiated from each of the individual subsets.
That is, we partition an option's start states into classes $\mathcal{C}$ such that $\prob{s^\prime | s, c} \approx \prob{s^\prime | c}, c \in \mathcal{C}$ (see Figure~2 in the supplementary material).
This can be practically achieved by clustering state transition samples based on effect states, and assigning each cluster to a partition. 
For each pair of partitions we then check whether their start states overlap significantly, and if so merge them, which accounts for probabilistic effects \citep{andersen17,konidaris18, ames18}.

Once we have partitioned subgoal options, we estimate the precondition and effect for each. 
Estimating the precondition is a classification problem, while the effect is one of density estimation.
Finally, for all valid combinations of effect distributions, we construct a  forward model by computing the probability that states drawn from their grounding lies within the learned precondition of each option, discarding rules with low probability of occurring. 

\section{Learning Portable Representations}

\begin{figure}[h!]

   		\centering   		 
   		\begin{subfigure}[b]{0.3\linewidth}
   		 	\centering
        	\includegraphics[height=30mm]{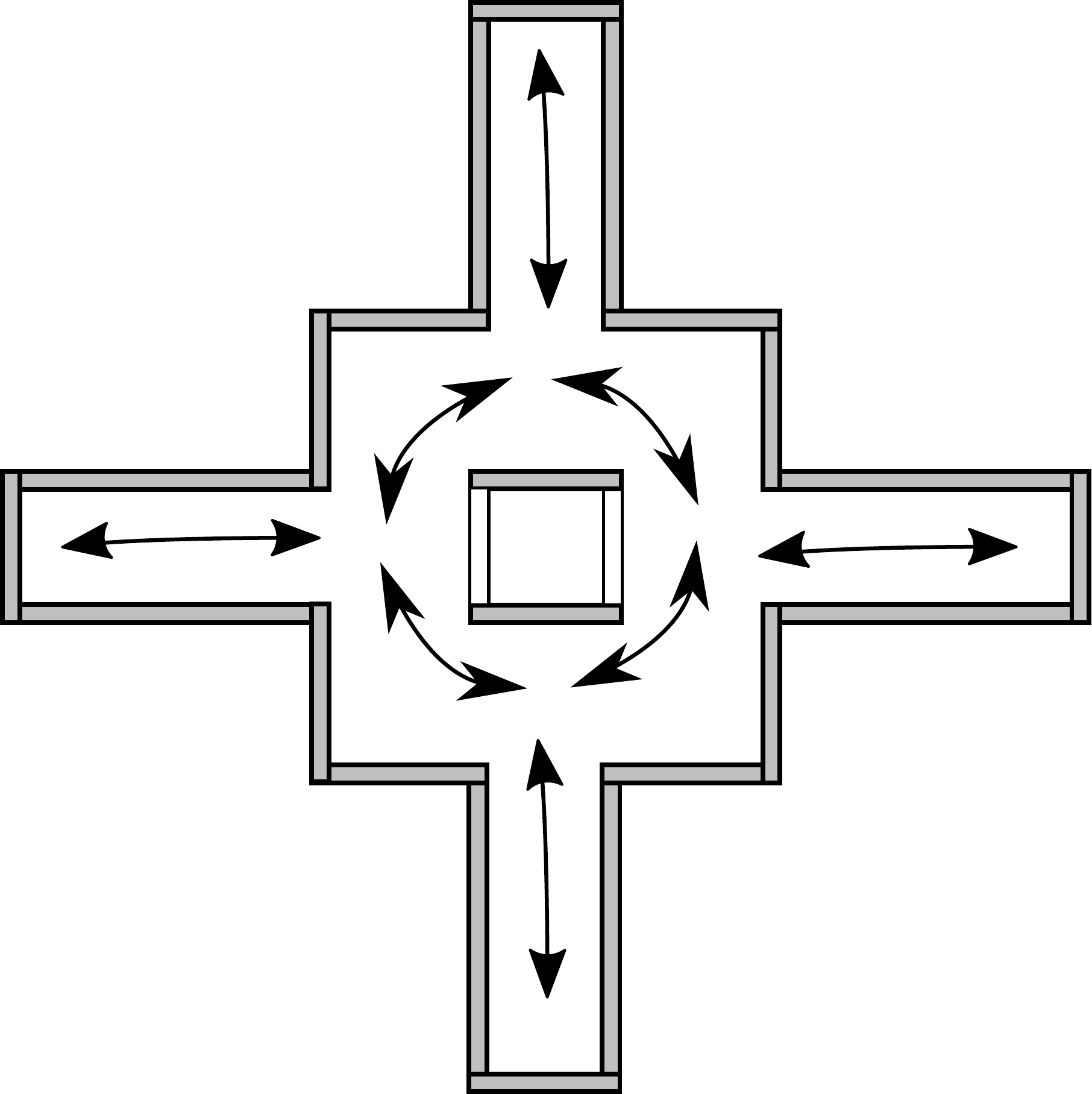}
        	\caption{}\label{fig:layout}
    	\end{subfigure}
   		 \begin{subfigure}[b]{0.2\linewidth}
   		 	\centering
        	\includegraphics[height=15mm]{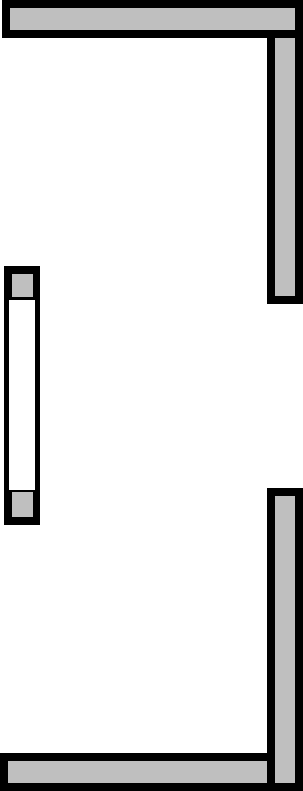}
        	\caption{}\label{fig:window-junction}
    	\end{subfigure}
        \begin{subfigure}[b]{0.2\linewidth}
   			\centering
        	\includegraphics[height=15mm]{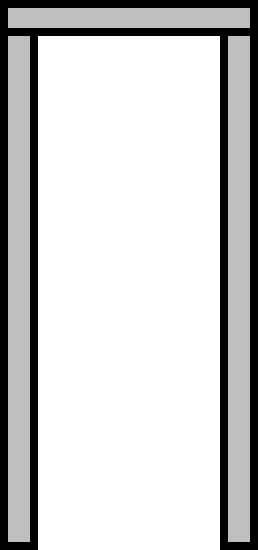}
        	\caption{} \label{fig:dead-end}
    	\end{subfigure}
   		\begin{subfigure}[b]{0.2\linewidth}
    		\centering
        	\includegraphics[width=15mm]{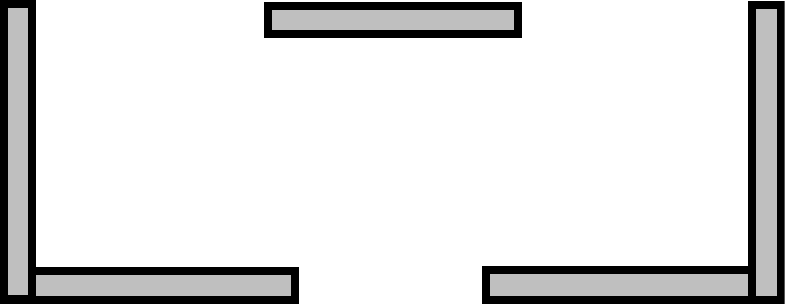}
        	\caption{}
        	\label{fig:wall-junction}
    \end{subfigure}

    \caption{(\subref{fig:layout}) A continuous navigation task where an agent navigates between different regions in $xy$-space. Walls are represented by grey lines, while the two white bars represent windows. Arrows describe the agent's options. (\subref{fig:window-junction}--\subref{fig:wall-junction}) Local egocentric observations. We name these \texttt{window-junction}, \texttt{dead-end} and \texttt{wall-junction} respectively.
    }\label{fig:toy}
\end{figure}

To aid in explanation, we use of a simple continuous task where a robot navigates the building illustrated in Figure~\ref{fig:layout}.
The problem space is the $xy$-coordinates of the robot, while we use an egocentric view of the environment (nearby walls and windows) around the agent for transfer. 
These observations are illustrated in Figure~\ref{fig:toy}.


The robot is equipped with options to move between different regions of the building, halting when it reaches the start or end of a corridor.
It possesses the following four options: (a) \texttt{Clockwise} and \texttt{Anticlockwise}, which move the agent in a clockwise or anticlockwise direction respectively, (b) \texttt{Outward}, which moves the agent down a corridor away from the centre of the building, and (c) \texttt{Inward}, which moves it towards the centre.

We could adopt the approach of \citet{konidaris18} to learn an abstract representation using transition data in $\mathcal{S}$.
However, that procedure generates symbols that are distributions over $xy$-coordinates, and are thus tied directly to the particular problem configuration. 
If we were to simply translate the environment along the plane, the $xy$-coordinates would be completely different, and our learned representation would be useless.

To overcome that limitation, we propose learning a symbolic representation over $\mathcal{D}$, instead of $\mathcal{S}$. 
Transfer can be achieved in this manner
because $\mathcal{D}$ remains consistent both within the same SMDP and across SMDPs, even if the state space or transition function do not.

Given only data produced by sensors, the agent proceeds to learn an abstract representation, identifying three portable symbols, which are exactly those illustrated by Figure~\ref{fig:toy}. 
The learned rules are listed in Table~\ref{tab:opts}, where it is clear that na\"ively considering egocentric observations alone is insufficient for planning purposes: the agent does not possess an option with probabilistic  outcomes,  but the \texttt{Inward} option appears to have probabilistic effects due to aliasing.

\begin{table*}[h!]
\centering
\caption{A list of the six subgoal options, specifying their preconditions and effects in agent space.} \label{tab:opts}

\begin{tabularx}{\linewidth}{lll}
\toprule
\textbf{Option} &
\textbf{Precondition} &
\textbf{Effect} \\
\midrule

\texttt{Clockwise1} & \texttt{wall-junction} & \texttt{window-junction} \\
\texttt{Clockwise2} & \texttt{window-junction} & \texttt{wall-junction} \\

\texttt{Anticlockwise1} & \texttt{wall-junction} & \texttt{window-junction} \\
\texttt{Anticlockwise2} & \texttt{window-junction} & \texttt{wall-junction} \\

\texttt{Outward} & \texttt{wall-junction} $\lor$ \texttt{window-junction} & \texttt{dead-end}  \\
\texttt{Inward} & \texttt{dead-end} & $
\begin{cases}
\text{\texttt{window-junction} w.p. }0.5\\
\text{\texttt{wall-junction} w.p. }0.5
\end{cases}
$ \\

\bottomrule
\end{tabularx}
\end{table*}

A further challenge appears when the goal of the task is defined in $\mathcal{S}$.
If we have goal $\mathcal{G} \subseteq \mathcal{S}$, then given information from $\mathcal{D}$, \textit{we cannot determine whether we have achieved the goal}.
This follows from the fact that the egocentric observations are lossy---two states $s, t \in \mathcal{S}$ may produce the same egocentric observation $d$, but if 
$s \in \mathcal{G}$ and $t \notin \mathcal{G}$, the knowledge of $d$ alone is insufficient to determine whether we have entered a state in $\mathcal{G}$.
We therefore require additional information to disambiguate such situations, allowing us to map from egocentric observations back into $\mathcal{S}$. 

\begin{figure}[htp]
\centering
\includegraphics[width=0.25\linewidth]{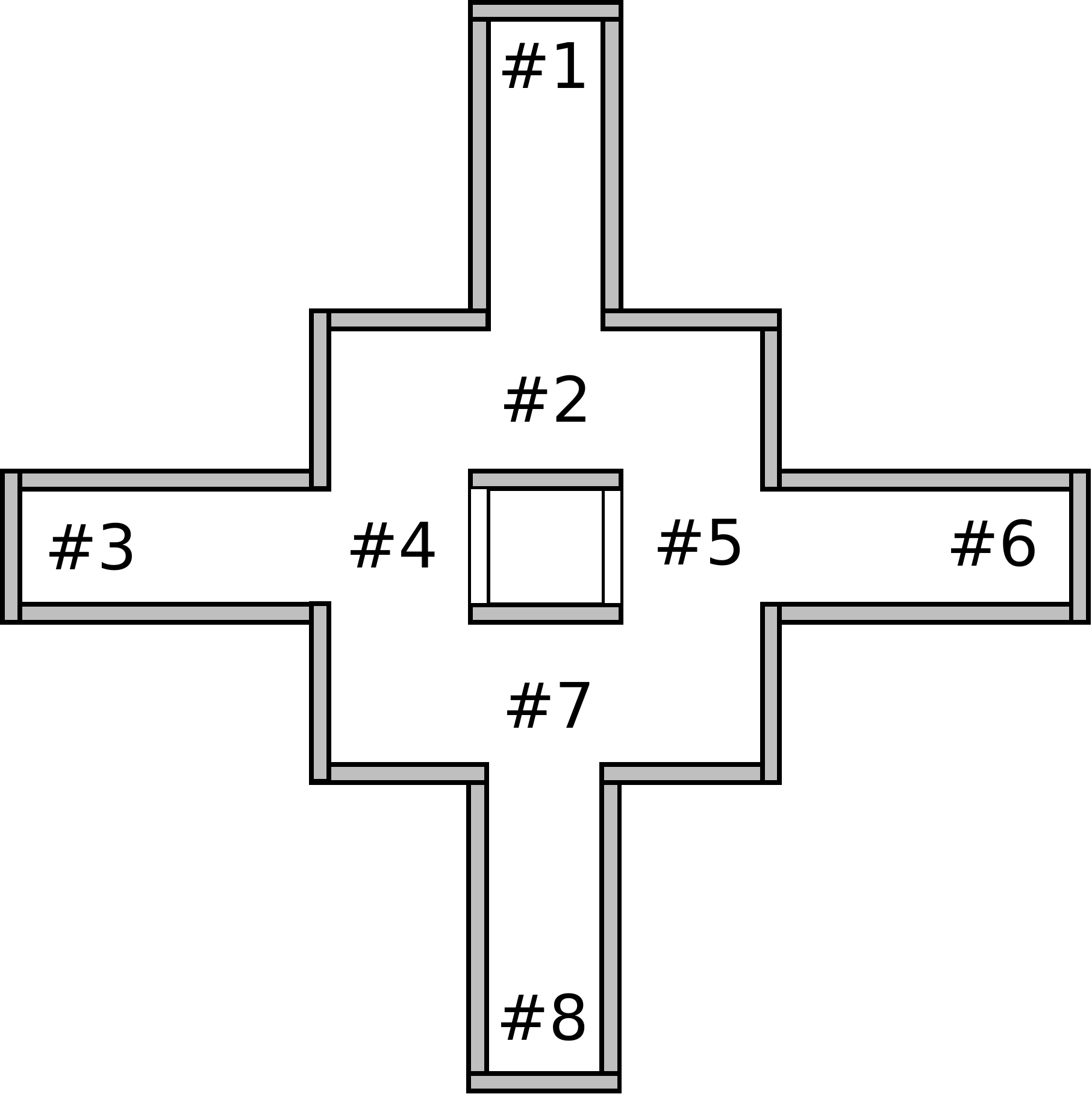}
\caption{Each number refers to the initiation set of an option partitioned in problem space. For readability, we merge identical partitions. For instance, \texttt{\#2} refers to the initation sets of a single problem space partition of \texttt{Outward}, \texttt{Clockwise} and \texttt{Anticlockwise}.} \label{fig:partitions}
\end{figure}

We can accomplish this by partitioning our portable options based on their effects in $\mathcal{S}$. 
This necessitates having access to both state and and egocentric observations. 
Recall that options are partitioned to ensure the subgoal property holds, and so each partition defines its own unique image distribution. 
If we label each problem-space partition, then each label refers to a unique distribution in $\mathcal{S}$ and is sufficient for disambiguating our egocentric symbols. 
Figure~\ref{fig:partitions} annotates the domain with labels according to their problem-space partitions.
Note that the partition numbers are completely arbitrary.

%
%
%
%
%

Generating agent-space symbols results in lifted symbols such as \texttt{dead-end(X)}, where \texttt{dead-end} is the name for a distribution over $\mathcal{D}$, and \texttt{X} is a partition number that must be determined on a per-task basis.
Note that the \textit{only} time problem-specific information is required is to determine the values of \texttt{X}, which grounds the portable symbol in the current task.


The following results shows that the combination of agent-space symbols with problem-space partition numbers provides a sufficient symbolic vocabulary for planning. (The proof is given in the supplementary material.) 

%
%
%
%
%
%

\begin{thm1} \label{thm}
 The ability to represent the preconditions and image of each option in egocentric space, together with goal $\mathcal{G}$'s precondition in problem space and partitioning in $\mathcal{S}$, is sufficient for determining the probability of being able to execute any probabilistic plan $p$ from starting distribution $Z$.
\end{thm1}

\section{Generating a Task-Specific Model}

\begin{figure}[h!]
\centering
\tikzstyle{block} = [rectangle, draw, text width=6em, text centered, rounded corners, minimum height=3em]
\tikzstyle{line} = [draw, -latex']

\scalebox{0.8}{%
\begin{tikzpicture}[node distance = 2cm, auto]
    
	\node (init)[align=left] {Given transition data \\ collected by executing \\ options};
    \node [block, right = 0.75cm of init, very thick, fill=colour] (partition) {Partition into subgoal options};
    \node [block, right = 0.75cm of partition, very thick, fill=colour] (estimate) {Estimate preconditions and effects};
    \node [block, right = 0.75cm of estimate, very thick, fill=colour] (generate) {Generate abstract forward model};
    \path [line, very thick] (init) -- (partition);
    \path [line, very thick] (partition) -- (estimate);
    \path [line, very thick] (estimate) -- (generate);    
    
    \node [block, below = 0.75cm of generate, very thick, fill=colour2] (partition2) {Partition options based on effects in $\mathcal{S}$};   
    \node [block, left = 0.75cm of partition2, very thick, text width=3cm, fill=colour2] (link) {Learn transitions between partition labels under each option};
    \node [block, left = 0.75cm of link, very thick, text width=3.5cm, fill=colour2] (ground) {Ground portable rules using partition labels for preconditions and effects.};
    \path [line, very thick] (generate) -- (partition2);
    \path [line, very thick] (partition2) -- (link);
    \path [line, very thick] (link) -- (ground);
\end{tikzpicture}
}

\caption{The full process of learning portable representations from data. Red nodes are learned using egocentric data from all previously encountered tasks, while 
 green nodes use problem-space data from the current task only.} \label{fig:process2}
\end{figure}
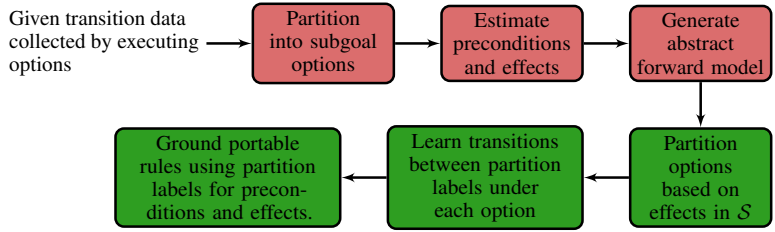

Our approach can be viewed as a two-step process.
The first phase learns portable symbolic rules using egocentric transition data from possibly several tasks, while the second phase uses problem-space transitions from the current task to partition options in $\mathcal{S}$.
The partition labels are then used as parameters to ground the previously-learned portable rules in the current task.
We use these labels to learn \textit{linking functions} that connect precondition and effect parameters.
For example, when the parameter of \texttt{Anticlockwise2} is \texttt{\#5}, then its effect should take parameter \texttt{\#2}. 
Figure~\ref{fig:process2} illustrates this grounding process.

These linking functions are learned by simply executing options and recording the start and end partition labels of each transition.  
We use a simple count-based approach that, for each option, records the fraction of transitions from one partition label to another. 
A more precise description of this approach is specified in the supplementary material.

A combination of portable rules and partition numbers reduces planning to a search over the space $\Sigma \times \mathbb{N}$, where $\Sigma$ is the set of generated symbols. Alternatively (and equivalently), we can generate either a factored MDP or a PPDDL representation \citep{younes04}.
To generate the latter, we use a \textit{function} named \texttt{partition} to store the current partition number and specify predicates for the three symbols derived in the previous sections: \texttt{window-junction}, \texttt{dead-end} and \texttt{wall-junction}.
The full domain description is provided in the supplementary material.

%

\section{Inter-Task Transfer}

In our example, it is not clear why one would want to learn portable symbolic representations---we perform symbol acquisition in $\mathcal{D}$ \textit{and} instantiate the rules for the given task, which requires more computation than directly learning symbols in $\mathcal{S}$. 
We now demonstrate the advantage of doing so by learning portable models of two different domains, both of which feature continuous state spaces and probabilistic transition dynamics. 

\subsection{Rod-and-Block} \label{sec:rod-block}

\begin{figure}[h!]
\centering
\includegraphics[width=0.6\linewidth]{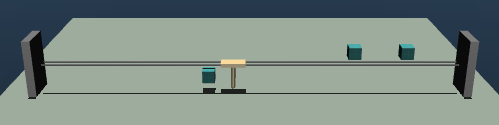}
\caption{The \textit{Rod-and-Block} domain. This particular task consists of three obstacles that prevent the rod from moving along the track when the rod is in either the upward or downward position. Different tasks are characterised by different block placements.} \label{fig:rod-block}
\end{figure}

We construct a domain we term \textit{Rod-and-Block} in which a rod is constrained to move along a track.
The rod can be rotated into an upward or downward position, and a number blocks are arranged to impede the rod's movement.
Two walls are also placed at either end of the track. 
One such task configuration is illustrated by Figure~\ref{fig:rod-block}.

The problem space consists of the rod's angle and its $x$ position along the track. 
Egocentric observations return the types of objects that are in close proximity to the rod, as well as its angle. 
In Figure~\ref{fig:rod-block}, for example, there is a block to the left of the rod, which has an angle of $\pi$.
The high-level options given to the agent are \texttt{GoLeft}, \texttt{GoRight}, \texttt{RotateUp}, and \texttt{RotateDown}. 
The first two translate the rod along the rail until it encounters a block or wall while maintaining its angle. 
The remaining options rotate the rod into an upwards or downwards position, provided it does not collide with an object.
These rotations can be done in both a clockwise and anti-clockwise direction.   

We learn a symbolic representation using egocentric transitions only, using the same procedure as prior work \citep{konidaris18}:
first, we collect agent-space transitions by interacting with the environment.
We partition the options in agent space using the DBSCAN clustering algorithm \citep{ester96} so that the subgoal property approximately holds.
This produces partitioned agent-space options.
Finally, we estimate the options' preconditions using a support vector machine with Platt scaling \citep{cortes95,platt99}, and use kernel density estimation \citep{rosenblatt56,parzen62} to model effect distributions.  

\begin{figure}[b!]%
    \centering
    \begin{subfigure}[t]{0.3\linewidth}
    	\centering
        \includegraphics[height=20mm]{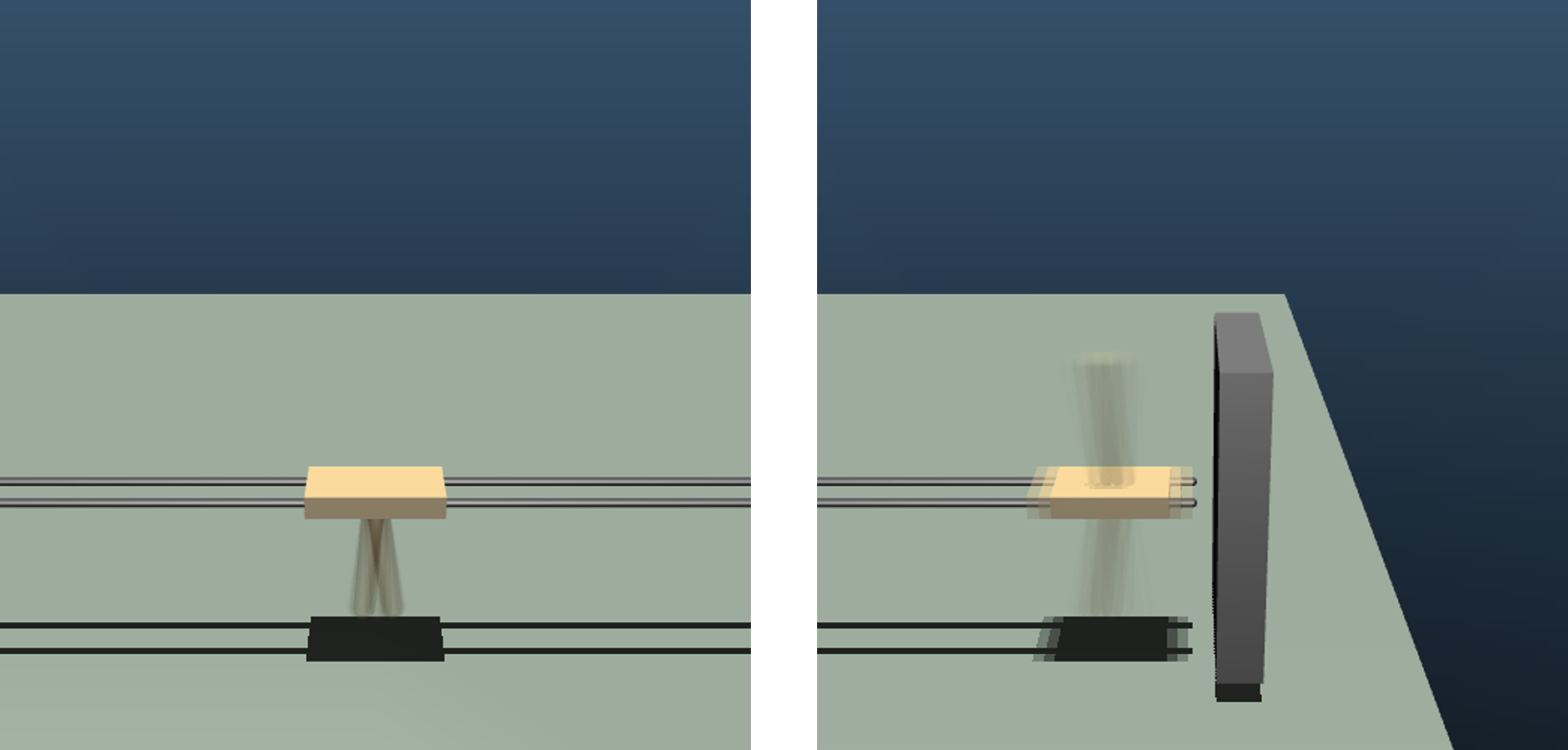}
        \caption{}
        \label{fig:rod-block_pre}
    \end{subfigure}
    \quad
            \begin{subfigure}[t]{0.3\linewidth}
    	\centering
        \includegraphics[height=20mm]{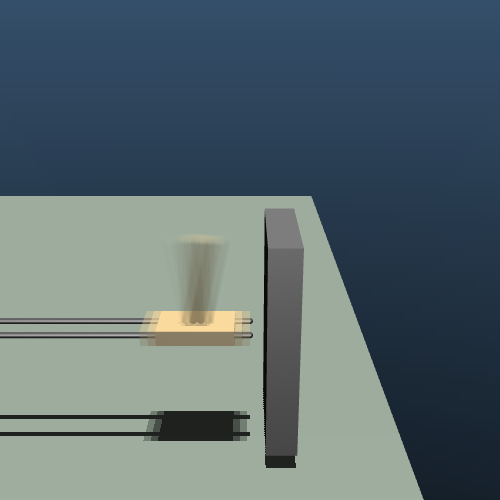}
        \caption{}
        \label{fig:rod-block_eff}
    \end{subfigure}     
        ~~
\begin{subfigure}[t]{0.3\linewidth}
    \centering
\begin{BVerbatim}[fontsize=\scriptsize]
(:action Up Clockwise_1
 :parameters()
 :precondition (and (symbol_18) 
               (symbol_11) 
               (notfailed))
 :effect (and (symbol_12) 
         (not symbol_18))
)
\end{BVerbatim}
\caption{}
\label{fig:pddl1}
\end{subfigure}      
    \caption{(\protect\subref{fig:rod-block_pre}) The precondition for \texttt{RotateUpClockwise1} operator, which states that in order to execute the option, the rod must be  left of a wall facing down. Note that the precondition is a conjunction of these two symbols---the first symbol is a distribution over the rod's angle only, while the second is independent of it. (\protect\subref{fig:rod-block_eff}) The effect of the option, with the rod adjacent to the wall in an upward position. (\protect\subref{fig:pddl1}) PDDL description of the above operator, which is used for planning.}%
            \label{fig:rod-block_symbols}
\end{figure}

The above procedure results in portable action rules, one of which is illustrated by Figure~\ref{fig:rod-block_symbols}. 
These rules can be reused for new tasks or configurations of the \textit{Rod-and-Block} domain---we need not relearn them when we encounter a new task,
though we can always use data from a new task to improve them.
More  portable rules are given in the supplementary material.

Once we have learned sufficiently accurate portable rules, the rules need only be instantiated for the given task by learning the linking between partitions.
This requires far fewer samples than classification and density estimation over the state space $\mathcal{S}$, which is required to learn a task-specific representation.

To illustrate this, we construct a set of ten tasks $\rho_1, \ldots, \rho_{10}$ by randomly selecting the number blocks, and then randomly positioning them along the track.  
Because tasks have different configurations, constructing a symbolic representation in problem space requires relearning a model of each task from scratch.
However, when constructing an egocentric representation, symbols learned in one task can immediately be used in subsequent tasks.
We gather $k$ transition samples from each task by executing options uniformly at random, and use these samples to build both task-specific and  egocentric (portable) models.

In order to evaluate a model's accuracy, we randomly select 100 goal states for each task, as well as the optimal plans for reaching each from some start state.
Each plan consists of two options, and we denote a single plan by the tuple $\langle s_1, o_1, s_2, o_2 \rangle$.
Let $\mathcal{M}^{\rho_i}_{k}$ be the model constructed for task $\rho_i$ using $k$  samples.
We calculate the likelihood of each optimal plan under the model: $\prob{ s_1 \in I_{o_1} | \mathcal{M}^{\rho_i}_{k}} \times \prob{ s^\prime \in I_{o_2} | \mathcal{M}^{\rho_i}_{k}}$, where $s^\prime \sim \text{Eff}(o_1)$. 
We build models using increasing numbers of samples, varying the number of samples in steps of 250, until the likelihood averaged over all plans is greater than some acceptable threshold (we use a value of $0.75$), at which point we continue to the next task.
The results are given by Figure~\ref{fig:rod-block-results}.

\subsection{Treasure Game}

We next applied our approach to the \textit{Treasure Game}, where an agent navigates a continuous maze in search of treasure. 
The domain contains ladders and doors which impede the agent. Some doors can be opened and closed with levers, while others require a key to unlock. 

The problem space consists of the $xy$-position of the agent, key and treasure, the angle of the levers (which determines whether a door is open) and the state of the lock. 
The egocentric space is a vector of length 9, the elements of which are the type of sprites in each of the nine directions around the agent, plus the ``bag'' of items possessed by the agent.
The agent possesses a number of high-level options, such as \texttt{GoLeft} and \texttt{DownLadder}. More details are given by  \citet{konidaris18}.  



\begin{figure}[h!]%
    \centering
    \begin{subfigure}[t]{0.3\linewidth}
    	\centering
        \includegraphics[height=33mm]{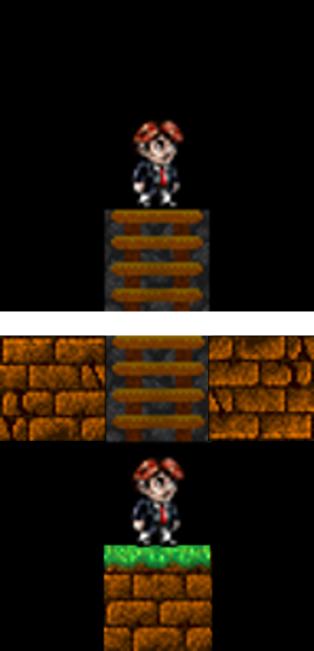}
        \caption{}
        \label{fig:pre_eff}
    \end{subfigure}    
    ~~
    \begin{subfigure}[t]{0.3\linewidth}
    	\centering
        \includegraphics[height=33mm]{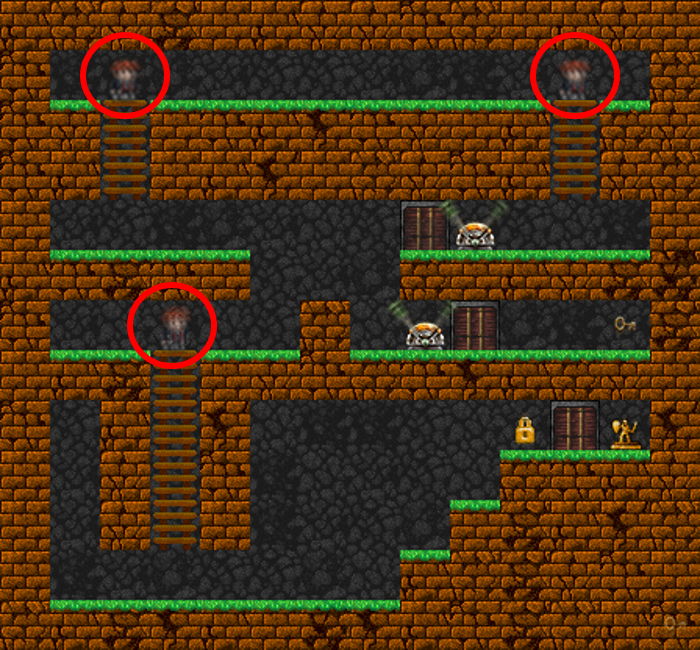}
        \caption{}
        \label{fig:labels}
    \end{subfigure}
    ~~
    \begin{subfigure}[t]{0.3\linewidth}
    \centering
\begin{BVerbatim}[fontsize=\scriptsize]

(:action DownLadder_1
 :parameters()
 :precondition (and (symbol_80) 
               (notfailed))
 :effect (and (symbol_622) 
              (not symbol_80))
)
\end{BVerbatim}
\caption{}
\label{fig:pddl2}
\end{subfigure}
    \caption{(\protect\subref{fig:pre_eff}) The precondition (top) and positive effect (bottom) for the \texttt{DownLadder} operator, which states that in order to execute the option, the agent must be standing above the ladder. The option results in the agent standing on the ground below it. The black spaces refer to unchanged low-level state variables. (\protect\subref{fig:labels}) Three problem-space partitions for the \texttt{DownLadder} operator. Each of the circled partitions is assigned a unique label and combined with the portable rule in (\protect\subref{fig:pre_eff}) to produce a grounded operator. (\protect\subref{fig:pddl2}) The PDDL representation of the operator specified in (\protect\subref{fig:pre_eff}).}%
    \label{fig:treasure_rule}
\end{figure}

\begin{figure}[h!]%
    \begin{subfigure}[t]{0.49\linewidth}
    \centering
		\includegraphics[width=0.8\linewidth]{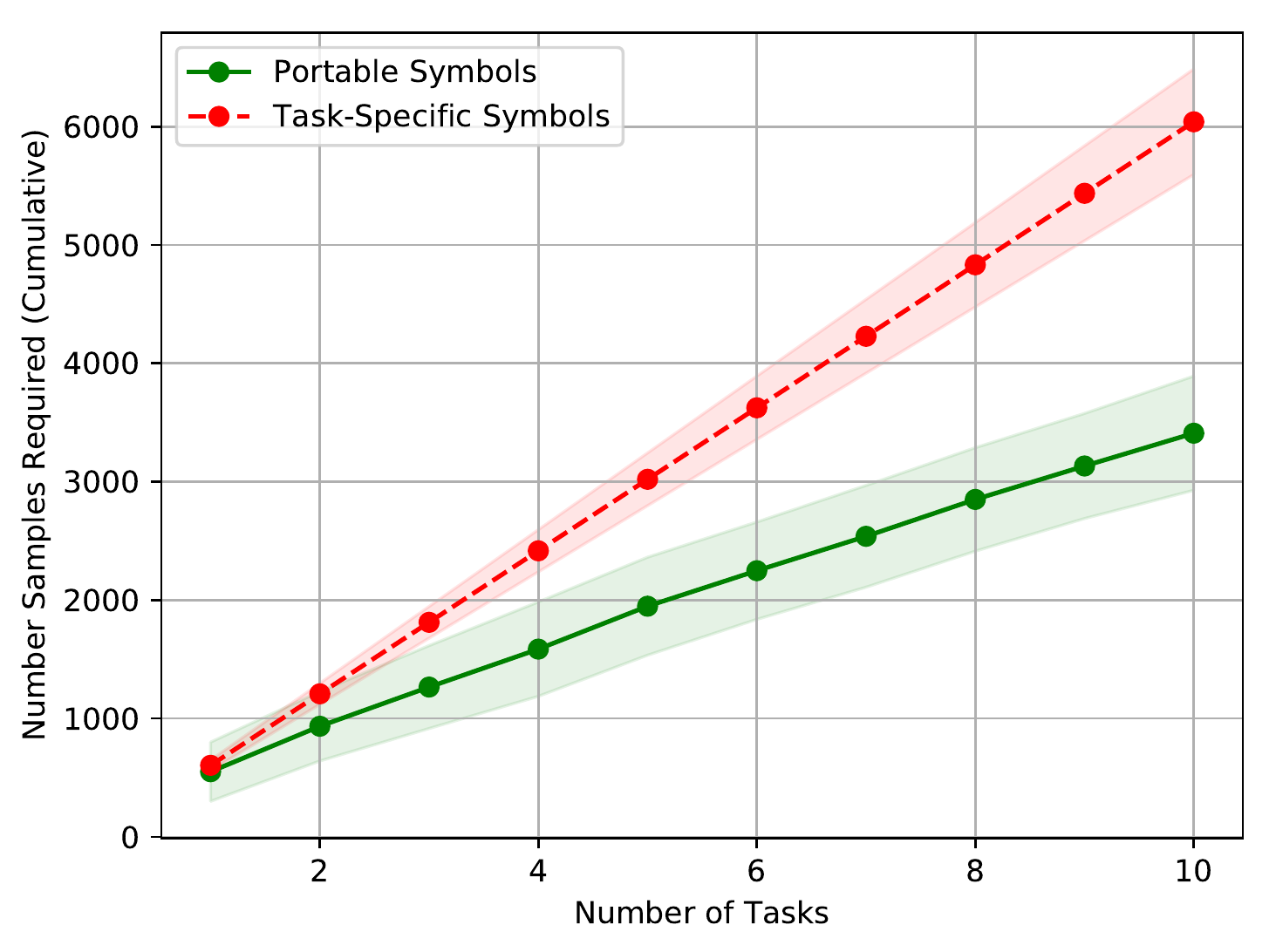}
		\caption{Results for the \textit{Rod-and-Block} domain.}
		\label{fig:rod-block-results}
    \end{subfigure}
	\quad 
	\begin{subfigure}[t]{0.49\linewidth}
    \centering
    	\includegraphics[width=0.8\linewidth]{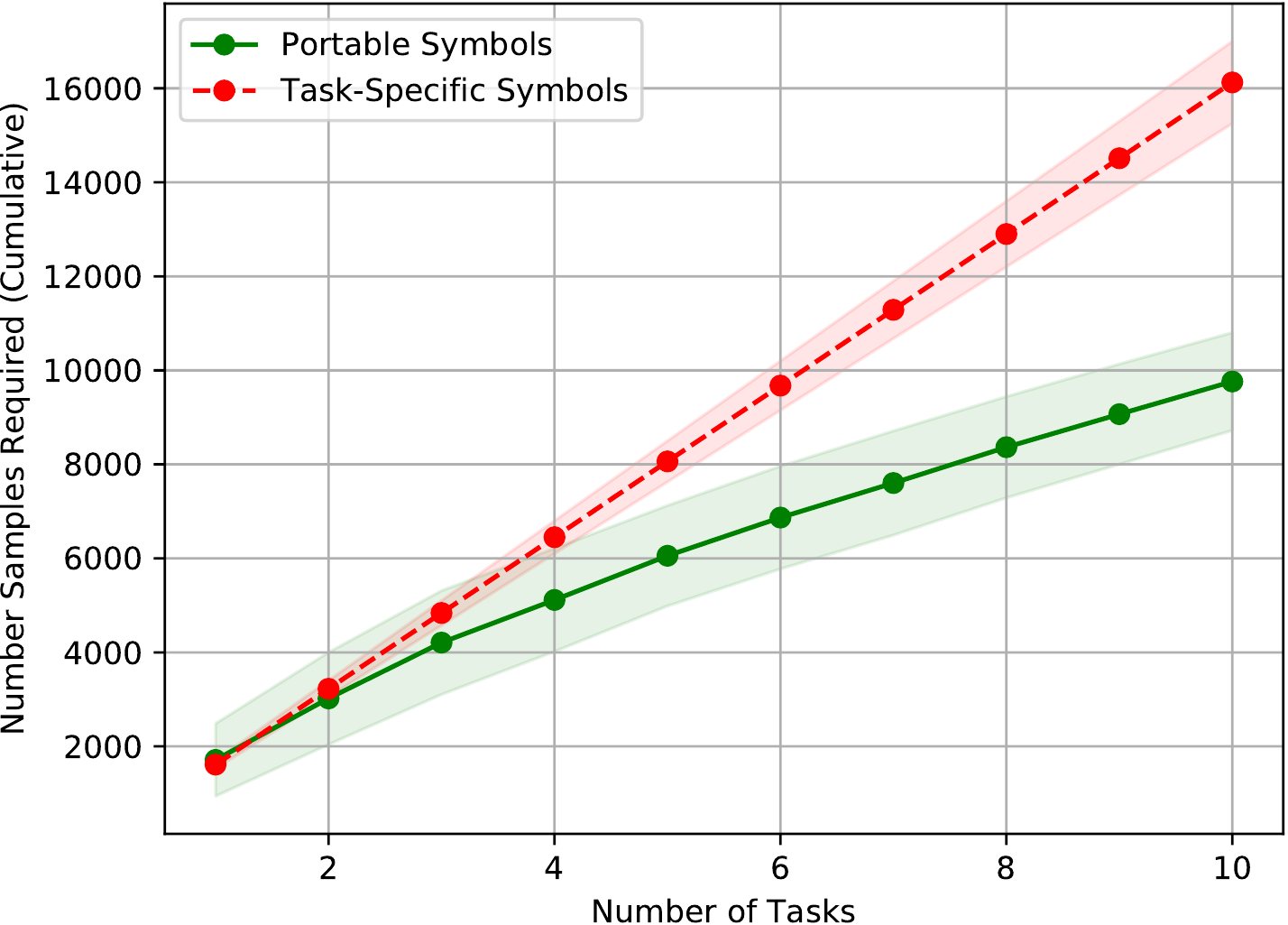}
    	\caption{Results for the \textit{Treasure Game} domain. }
		\label{fig:results}
    \end{subfigure}
    \caption{Cumulative number of samples required to learn sufficiently accurate models as a function of the number of tasks encountered. Results are averaged over 100 random permutations of the task order. Standard errors are specified by the shaded areas.}
\end{figure}

We construct a set of ten tasks $\rho_1, \ldots, \rho_{10}$ corresponding to different levels of the \textit{Treasure Game},\footnote{We made no effort to design tasks in a curriculum-like fashion. The levels are given in the supplementary material.} and learn portable models and test their sample efficiency as in Section~\ref{sec:rod-block}. 
An example of a portable action rule, as well as its problem-space partitioning, is given by Figure~\ref{fig:treasure_rule}, while the number of samples required to learn a good model of all 10 levels is given by Figure~\ref{fig:results}.

%

\subsection{Discussion}


Naturally, learning problem-space symbols results in a sample complexity that scales linearly with the number of tasks, since we must learn a model for each new task from scratch.
Conversely, by learning and reusing portable symbols, we can reduce the number of samples we require as we encounter more tasks, leading to a \textit{sublinear} increase.
The agent initially requires about $600$ samples to learn a task-specific model of each \textit{Rod-and-Block} configuration, but decreases to roughly $330$ after only two tasks. 
Similarly, $1600$ samples are initially needed for each level of the \textit{Treasure Game}, but only $900$ after four levels, and about $700$ after seven.

Intuitively, one might expect the number of samples to plateau as the agent observes more tasks. 
That we do not is a result of the exploration policy---the agent must observe all relevant partitions at least once, and selecting actions uniformly at random is naturally suboptimal.  
Nonetheless, we still require far fewer samples to learn the links between partitions than does learning a full model from scratch.
 
In both of our experiments, we construct a set of 10 domain configurations and then test our approach by sampling 100 goals for each, for a total of 1000 tasks per domain.
Our model-based approach learns 10 forward models, and then uses them to plan a sequence of actions to achieve each goal.
By contrast, a model-free approach \citep{jonschkowski15, higgins17, kirkpatrick17, finn17, debruin18} would be required to learn all 1000 policies, since every goal defines another unique SMDP that must be solved. 
 

%
%
%

\section{Related Work}

There has been some work in autonomously learning parameterised representations of skills, particularly in the field of relational reinforcement learning.
\citet{finney02}, \citet{pasula04} and \citet{zettlemoyer05}, for instance, learn operators that transfer across tasks.
However, the high-level symbolic vocabulary is given; we show how to learn it.
\citet{ames18} adopts a similar approach to \citet{konidaris18} to learn symbolic representations for parameterised actions.
However, the representation learned is fully propositional (even if the actions are not) and cannot be transferred across tasks. 


\textit{Relocatable action models} \citep{sherstov05,leffler07} assume states can be aggregated into ``types'' which determine the transition behaviour.  
State-independent representations of the outcomes from different types are learned and improve the learning rate in a single task. 
However, the mapping from lossy observations to states is provided to the agent, since learning this mapping is as hard as learning the full MDP.


More recently, \citet{zhang18} propose a method for constructing portable representations for planning. However, the mapping to abstract states is provided, and planning is restricted solely to the equivalent of an egocentric space.
Similarly, \citet{srinivas18} learn a goal-directed latent space in which planning can occur. 
However, the goal must be known up front and be expressible in the latent space. 
We do not compare to either, since both are unsuited to tasks with goals defined in problem space, and neither provide soundness guarantees.


%

\section{Summary}

We have introduced a framework for autonomously learning portable symbols given only data gathered from option execution, and showed that the addition of particular problem-specific information results in a representation that is provably sufficient for learning a sound representation for planning. 
This allows us to leverage experience in solving new unseen tasks---an important step towards creating adaptable, long-lived agents.

\bibliographystyle{unsrtnat}

\bibliography{portable_symbols}

\end{document}